\documentclass[10pt,journal,compsoc]{IEEEtran}

\usepackage{biblatex}
\usepackage[pdftex]{graphicx}
\usepackage{amsmath}
\usepackage{algorithmic}
\usepackage{array}
\usepackage[caption=false,font=footnotesize,labelfont=sf,textfont=sf]{subfig}
\usepackage{url}
\usepackage[pdftex]{thumbpdf}

\usepackage{color}
\usepackage{soul}

\newcommand{\citationneeded}[1][]{\textsuperscript{[citation needed]}}

\begin{document}

\title{Cerbero-7B: A Leap Forward in Language-Specific LLMs Through Enhanced Chat Corpus Generation and Evaluation}


\author{Federico~A.~Galatolo, Mario~G.C.A.~Cimino (\IEEEmembership{Member, IEEE})}

\markboth{Journal of \LaTeX\ Class Files,~Vol.~14, No.~8, August~2015}%
{Shell \MakeLowercase{\textit{et al.}}: Bare Advanced Demo of IEEEtran.cls for IEEE Computer Society Journals}

\IEEEtitleabstractindextext{%

\begin{abstract}
This study introduces a novel approach for generating high-quality, language-specific chat corpora using a self-chat mechanism. We combine a generator LLM for creating new samples and an embedder LLM to ensure diversity. A new Masked Language Modelling (MLM) model-based quality assessment metric is proposed for evaluating and filtering the corpora. Utilizing the \texttt{llama2-70b} as the generator and a multilingual sentence transformer as embedder, we generate an Italian chat corpus and refine the \textit{Fauno} corpus, which is based on translated English ChatGPT self-chat data. The refinement uses structural assertions and Natural Language Processing techniques. Both corpora undergo a comprehensive quality evaluation using the proposed MLM model-based quality metric. The Italian LLM fine-tuned with these corpora demonstrates significantly enhanced language comprehension and question-answering skills. The resultant model, \texttt{cerbero-7b}, establishes a new state-of-the-art for Italian LLMs. This approach marks a substantial advancement in the development of language-specific LLMs, with a special emphasis on augmenting corpora for underrepresented languages like Italian.
\end{abstract}

}

\maketitle

\IEEEdisplaynontitleabstractindextext
\IEEEpeerreviewmaketitle

\section{Introduction}\label{sec:introduction}

In the dynamic domain of Natural Language Processing (NLP), the enhancement of conversational models has become a pivotal area of research \cite{conv-llm-survey}. The advent of sophisticated models, notably ChatGPT and GPT-4, has demonstrated exceptional capabilities in emulating human-like discourse across a multitude of contexts \cite{instructgpt} \cite{gpt4}. Nonetheless, the exclusivity of APIs and the dearth of publicly accessible, high-caliber chat corpora have imposed significant constraints on the academic and development communities.

Advancing the pioneering methodologies delineated in the seminal works of the Fauno \cite{fauno2023italian} and Baize \cite{baize2023open} studies, this paper introduces a distinctive approach to the generation of a diversified and superior-quality corpus through the medium of Large Language Model (LLM) self-chat. Our method is characterized by its exclusive reliance on open-source software and models, while striving to attain, if not surpass, the corpus quality derived from proprietary self-chat paradigms, such as those employed by ChatGPT.

This research focuses on the refinement of an Italian LLM. We engaged with two distinct Italian corpora: Fauno, which, despite its considerable size, offers lower quality \cite{fauno2023italian}, and OASST \cite{oasst}, which, though limited in scale, provides high-quality, human-curated chat data. Our goal is to leverage these repositories to fortify the linguistic proficiency of an Italian LLM.

Confronting the inherent qualitative deficiencies of the Fauno corpus, we advocate a novel technique that incorporates structural assertions and rule-based NLP for refinement. Concurrently, we delineate a novel method for generating a new chat corpus via a self-chat process augmented by a sentence embedding transformer.

To assess the quality of the corpus, we employ novel a BERT-based method to evaluate and compare the original, filtered, and generated corpora. This evaluative process informs the fine-tuning phase, wherein the corpora are employed to fine-tune a state-of-the-art LLM.

Our research culminates in a comprehensive evaluation, wherein we benchmark the resulting model's capabilities in linguistic comprehension and question-answering tasks, comparing its performance with other leading Italian Large Language Models. In a stride towards promoting collaborative research and development, we have made the most effective fine-tuned model publicly available under the name \texttt{cerbero-7b}.

The choice of this nomenclature, which follows the common practice of naming fine-tuned models from animal or mythological creatures, is inspired by Cerberus, the mythological three-headed dog tasked with guarding the gates of the Underworld in ancient Greek mythology. This metaphor aptly represents the tripartite foundation of our model: (i) its base model, the \texttt{mistral-7b}, which provides a robust and advanced starting framework; (ii) the innovative approach in corpus generation and evaluation developed in this study; and (iii) the commitment to open-source development, as demonstrated by the adoption of the Apache 2.0 license, fostering an environment of accessibility and collaboration within the AI community.

\section{Related Works}

The evolution of large language models (LLMs) has significantly impacted the field of Natural Language Processing (NLP), particularly in the context of chat-based applications. Since the inception of transformer-based models like BERT \cite{bert} and GPT \cite{radford2018gpt}, the approach to NLP tasks has been revolutionized, achieving state-of-the-art performance across various benchmarks \cite{zhang2019dialogpt}.

Recent methodologies aim to leverage the vast knowledge embedded in these pre-trained models through fine-tuning techniques that emulate human judgments for specific tasks and evaluation criteria, showing promise in automatic dataset generation for NLP. Concurrently, the field has seen a surge in the use of these models to automatically generate corpora for NLP tasks. \cite{baize2023open}

One of the most prominent contributions to this domain has been the development of open-source alternatives to proprietary models. Models like Stanford Alpaca and Vicuna have almost reached the performance of models like ChatGPT and GPT-4 by fine-tuning on corpora collected from various sources, including instruction-learning formats and public dialogue data \cite{openai2023chatgpt, alpaca, vicuna2023}.

In parallel to advancements in dataset generation, there has been significant progress in fine-tuning LLMs, especially in resource-constrained scenarios. Methods like LOw-Memory Optimization (LOMO), which fuses gradient computation and parameter update in one step, have emerged as parameter-efficient alternatives to traditional full-parameter fine-tuning \cite{lv2023lomo}. However, the challenge remains in fine-tuning the full parameters of LLMs, particularly for models with tens to hundreds of billions of parameters, which often necessitate considerable computational resources \cite{wei2022emergent}.

Methods like LOw-Memory Optimization (LOMO), which fuses gradient computation and parameter update in one step, have emerged as parameter-efficient alternatives to traditional full-parameter fine-tuning. Additionally, Low-Rank Adaptation (LoRA) has been introduced to reduce the number of trainable parameters by injecting trainable rank decomposition matrices into each layer of the Transformer architecture, allowing for efficient fine-tuning of large models like GPT-3 \cite{hu2022lora}.

The open-source model Baize utilizes parameter-efficient tuning on self-chat data to enhance LLaMA, an open-source large language model, demonstrating impressive multi-turn dialogue performance and accessibility for research on a single GPU \cite{baize2023open}.

In the context of Italian language processing, significant strides have been made with the development of Italian-specific LLMs. GePpeTto represents the first generative language model for Italian, trained on a corpus comprising Italian Wikipedia and the ItWac corpus, using GPT-2 as a blueprint and amounting to almost 13GB of text \cite{geppetto2020}. Its training utilized four Tesla T4 GPUs, with a model size corresponding to GPT-2 small, including 12 layers and 117 million parameters.

Building on this, the IT5 model family has been introduced as encoder-decoder transformer models pre-trained specifically on Italian. Utilizing a cleaned version of the Italian mC4 corpus, IT5 models have been shown to consistently outperform multilingual counterparts, setting new benchmarks for most Italian language generation tasks \cite{it5}. This advancement solidifies the impact of monolingual models on language-specific NLP performance.

The Fauno model pushes the envelope further as the first and largest open-source Italian conversational LLM. Aimed at democratizing LLM studies in Italian, Fauno was fine-tuned on a diverse range of corpora including general QA, computer science, and medical questions, to offer a capable conversational AI in Italian \cite{fauno2023italian}. 

Complementing these Italian-specific developments, the multilingual SBERT, particularly the \textit{distiluse-base-multilingual-cased} variant, leverages knowledge distillation to extend monolingual sentence embedding models to multiple languages. This method maps translated sentences to the same vector space as the original, enabling the creation of multilingual embeddings from previously monolingual models. This approach ensures desired properties for the vector space with lower hardware requirements, demonstrating effectiveness across 50+ languages \cite{sbert}.

\section{Methodology}

\subsection{Fauno}

In this methodology subsection, we delineate the procedures for sanitizing and preprocessing the \textit{Fauno} corpus—a numerous yet qualitatively deficient Italian corpus, derived from English ChatGPT self-chat data. Our aim is to reinforce the corpus's structural integrity, linguistic coherence, and readiness for the subsequent fine-tuning of an Italian Language Model (LLM). We expound upon these procedures below.

Originally, the \textit{Fauno} corpus was formatted as text, with the first line as the system prompt, followed by alternating responses from human and AI interlocutors, marked with the tags \texttt{[|Umano|]} and \texttt{[|AI|]}, respectively. The translation from English introduced inconsistencies, including erroneous tags like \texttt{[| Human |]} or misaligned brackets, and omitted annotations. To instill uniformity, we employed regular expressions for reformatting.

Our primary focus was to maintain the natural conversation flow, with the AI providing responses before the human participants. To achieve this, we undertook a two-step process. First, we removed 73 conversations that were empty or devoid of meaningful content. These empty conversations added no value to our corpus and were, therefore, excluded. 

In the second step, we meticulously reviewed and identified 225 conversations that deviated from the desired flow sequence of AI followed by Human. These deviations disrupted the desired conversational structure, and as a result, they were also excluded from the corpus.

In a further step, we computed the hash of each message using the \texttt{xxhash} hash function. This allowed us to identify and remove conversations with more than 50\% of messages identical to others in the corpus. Consequently, we removed an additional 2368 conversations, further refining the data quality.

The linguistic analysis utilized the \textit{NLTK} \textit{Punkt} tokenizer for sentence segmentation \cite{nltk} and \textit{Lingua-py} for language detection \cite{linguapy}. We discovered 67,517 English messages within the conversations, 50,245 of which were untranslated system prompts.

All of the system prompts lacked meaningful content as they were consistently identical or had limited variations. They essentially served as standardized triggers for the AI to generate responses and did not contribute any valuable linguistic diversity or information to the corpus. Consequently, we removed the system prompts from the cleaned corpus

In the final step, a zero-shot classifier, designed to differentiate between text and code and built using the \textit{llama2-70b-chat} model, was deployed to analyze the remaining 17,272 detected English messages. This classification process revealed that the vast majority of the detected ``English'' messages actually contained code and were not genuine English text. Upon closer examination, we found that all of the remaining messages that appeared to be in English were, in fact, short sentences containing English names but were written in Italian. These instances turned out to be false positives.

\begin{figure}
    \centering
    \includegraphics[width=0.8\linewidth]{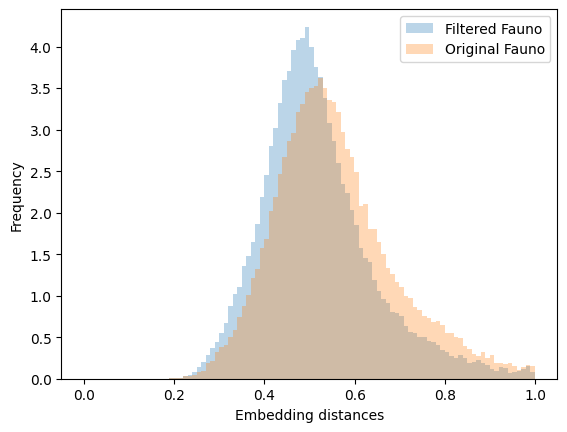}
    \caption{Sentence embedding distance distribution over the original and filtered \textit{Fauno} corpora}
    \label{fig:filtered-fauno}
\end{figure}

For quantitative validation of the cleaning process, we utilized the \textit{distiluse-base-multilingual-cased-v1} sentence transformer model. By embedding each message and storing these within a vector store, we identified the ten nearest neighbors for each sentence, discounting identical embeddings. A histogram of these distances, depicted in Figure \ref{fig:filtered-fauno}, illustrates the enhanced cohesiveness of our refined corpus. This approach provides insights into the distribution of differences within the corpus. A more diverse corpus is deemed superior, emphasizing the importance of our cleaning methodology in enhancing the overall corpus quality.

\subsection{OASST}

The OASST dataset is a curated corpus that captures a wide range of linguistic nuances through its conversation trees \cite{oasst}. Each tree in the corpus begins with a root message, branches out into multiple dialogues, and culminates in leaf nodes, signaling the conclusion of an exchange. This hierarchical structure mirrors the complexity and layered nature of real-life conversations, offering a rich resource for analyzing linguistic patterns and conversational dynamics.

Specifically, the Italian portion of the OASST corpus, which is central to our analysis, comprises 111 conversation trees encompassing a total of 554 messages. Our methodology involved a comprehensive exploration of the conversation trees in the Italian subset. Starting from the root messages, we traced each dialogue branch to its corresponding leaf node.
This process allowed us to identify and extract 358 seed conversations. These seed conversations were then utilized as a foundation for further analysis and the generation of self-chat scenarios.

\subsection{Italian Chat Corpus Generation}

The generation of a new Italian chat corpus was conducted by initially sampling ten seed conversations from the OASST corpus, specifically from its Italian section. These conversations were then employed as the foundation for further message generation using the \textit{llama2-70b model} \cite{llama2}. This process utilized the model in a self-chat configuration, aiming to extend each conversation to a predetermined target length, randomly sampled for each conversation from 4 to 10 messages.

For every newly generated chat message, we computed its embeddings using the sentence transformer \texttt{distiluse-base-multilingual-case} \cite{sbert}. This step transformed the messages into high-dimensional vectors, facilitating their comparison for uniqueness and diversity. These embeddings were crucial for the subsequent filtering process.

\begin{figure}[ht]
    \centering
    \includegraphics[width=\linewidth]{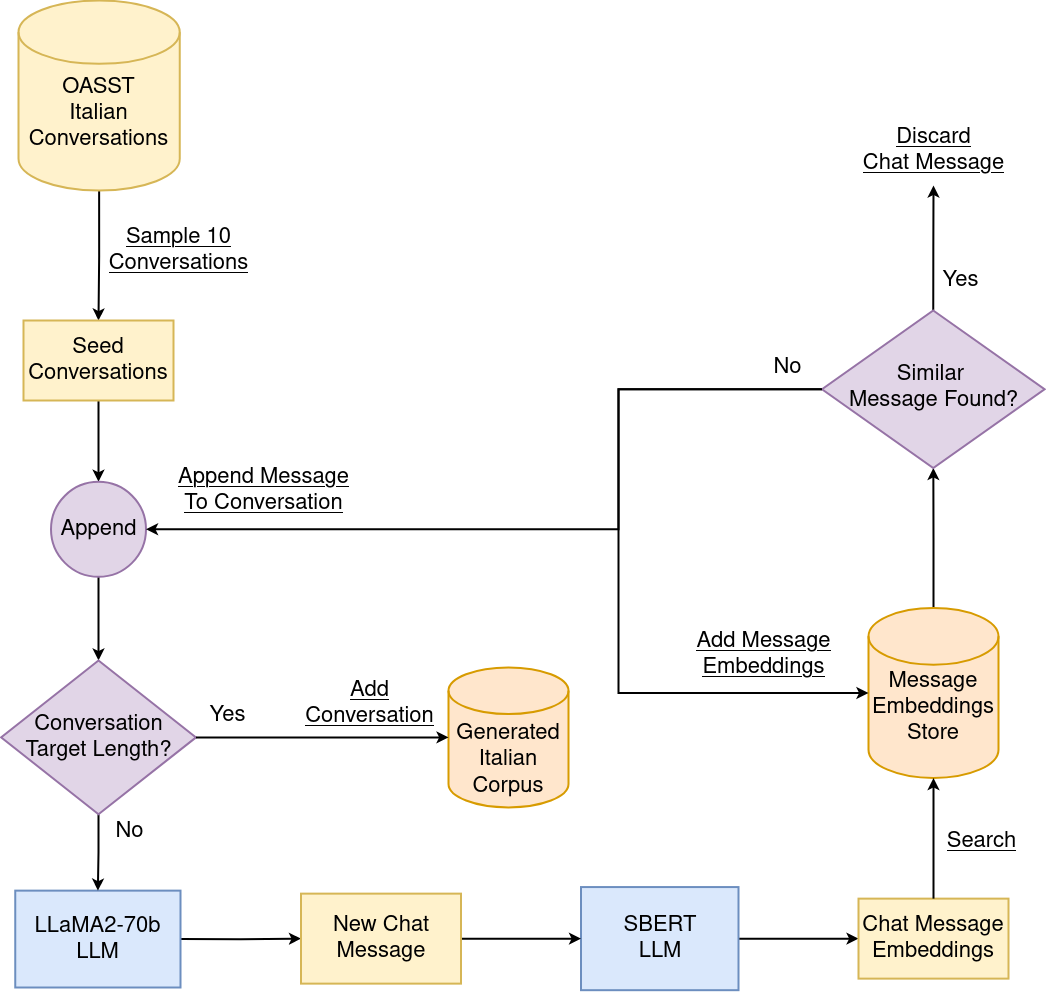}
    \caption{Flowchart of the Italian Chat Corpus Generation process}
    \label{fig:generation}
\end{figure}

A crucial aspect of our methodology was the introduction of a diversity metric based on cosine similarity measures. A Vector Store database, initially populated with every OASST message, was queried with each new message embedding to check for similarity with existing messages. Messages displaying a cosine similarity above 0.9 with any vector in the database were excluded. This diversity filter ensured that only unique and varied linguistic elements were added to the corpus, thereby enriching the corpus with a wide spectrum of linguistic expressions.

Messages that successfully passed the diversity filter were appended to their respective conversations. Concurrently, their embeddings were added to the Vector Store database. This iterative process of conversation and database enrichment led to the formation of a dynamic and contextually rich Italian chat corpus.

The entire generation process is depicted in Figure \ref{fig:generation}: cyan blocks in the flowchart represent the LLM models employed, while purple blocks denote the control logic guiding the message generation and filtering. Yellow blocks symbolize the objects in play, namely the messages and their embeddings. Finally, orange blocks represent the databases involved in the process, including both the Vector Store and the generated corpus. This systematic approach was instrumental in achieving our objective of expanding the Italian chat corpus with high-quality, diverse conversational content.

\subsection{Quality Assesment}

In this section, we elaborate on our methodology for assessing the quality of the generated corpus through the utilization of BERT-based language models (LLMs). Our approach seeks to evaluate the linguistic coherence and contextual understanding of the corpus sentences, ultimately contributing to the enhancement of the performance of the fine-tuned LLM on downstream tasks.
To initiate the quality assessment process, each sentence in the corpus is subjected to tokenization. Subsequently, each token within the sentence is individually masked, one at a time. This meticulous token-wise masking enables a granular examination of the language model's predictive capabilities when faced with the absence of specific tokens.
The probability distribution over the masked tokens is computed using an Italian BERT model. For every masked token $T_i$, the model generates a probability distribution conditioned on the entire sentence, excluding the masked token $p(T_i\,|T_j) \; \forall j \ne i$

\begin{figure}
    \centering
    \includegraphics[width=0.85\linewidth]{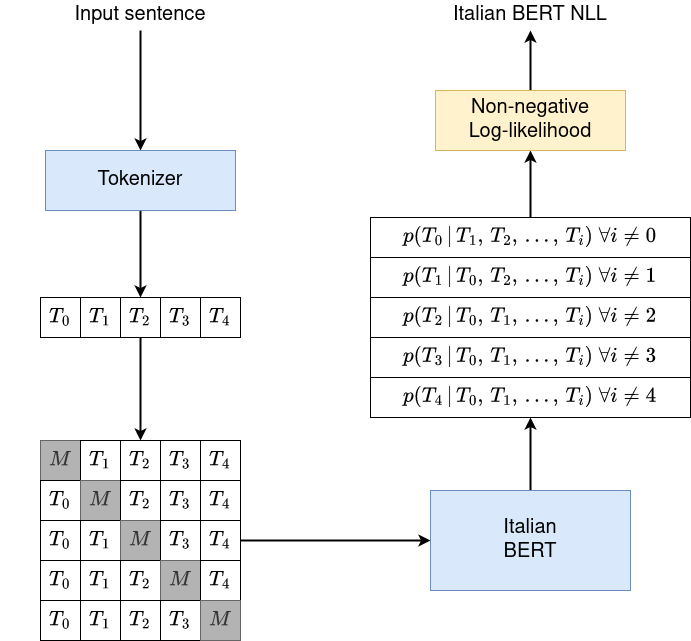}
    \caption{Illustration of the Proposed Quality Metric Evaluation Process}
    \label{fig:evaulation}
\end{figure}

This step aims to capture the model's prediction of the masked token given the contextual information provided by the rest of the tokens in the sentence.
The Non-Negative Log-Likelihood (NLL) is employed as a metric to quantify the quality of the sentence. For each masked token in the sentence, the NLL is computed based on the actual token observed in the corpus. Mathematically, the NLL is defined as:

\begin{equation}
NLL = \frac{1}{N}\sum\limits_{i=0}^{N}-log(p(T_i\,|T_j) \; \forall j \ne i
\end{equation}
Here, the negative logarithm of the predicted probability is taken, and the mean NLL across all masked tokens within the sentence is calculated. This process is repeated for each sentence in the corpus.
The rationale behind utilizing NLL as a quality assessment metric lies in its sensitivity to the predictability of the masked tokens. Higher NLL scores indicate greater uncertainty and randomness in the language model's predictions, suggesting potential issues such as sentence malformation or lack of coherence. Conversely, lower NLL scores signify more predictable and contextually aligned predictions, indicative of well-structured and semantically sound sentences. The evaluation process is depicted in Figure \ref{fig:evaulation}

The computed NLL scores are utilized to evaluate and compare the quality of the original corpus, the filtered corpus, and the generated one. This comparative analysis serves as a crucial step in gauging the efficacy of our corpus generation methodology and its impact on language model performance. 

\begin{figure}
    \centering
    \includegraphics[width=0.8\linewidth]{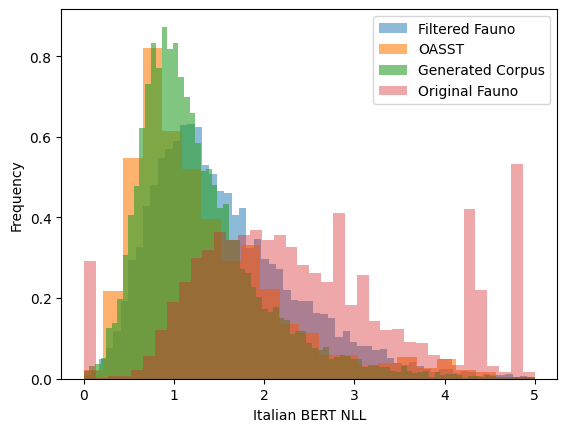}
    \caption{NLL Quality Metric distribution for different chat corpora}
    \label{fig:bert-histogram}
\end{figure}

The histogram, depicted in Figure \ref{fig:bert-histogram}, provides a comparative analysis of the Non-Negative Log-Likelihood (NLL) quality metric distribution across different chat corpora. From this histogram, we observe that the OASST dataset exhibits the lowest mean NLL score and a relatively narrow distribution. This indicates a high degree of linguistic coherence and contextual consistency within the OASST corpus, reflecting its curated nature. In contrast, the Generated corpus demonstrates a distribution pattern strikingly similar to that of OASST, with a marginally higher mean but comparable dispersion, implying a successful emulation of OASST's quality through our corpus generation methodology.

The Filtered Fauno corpus, while showing a broader dispersion and a higher mean NLL score than OASST and Generated, still maintains proximity to these datasets. This suggests that, despite its refinement, the Filtered Fauno corpus possesses a somewhat varied range of linguistic coherence, likely due to its origins from a translation of English ChatGPT self-chat data. The broader dispersion signifies a more heterogeneous composition in terms of language quality, yet still denotes significant improvement from its original form.

In stark contrast, the Original Fauno corpus is characterized by a substantially higher mean and a considerably broader dispersion in its NLL score distribution. This pattern is indicative of a noisy and less coherent dataset. The pronounced peaks in the histogram of the Original Fauno corpus further underscore its inconsistency and the varied quality of its linguistic content. Such a distribution profile highlights the substantial qualitative gap between the Original Fauno and the other corpora, underlining the critical need for the refinement and generation processes we have employed in this study.

\section{Experimental Setup}

This section provides an overview of the experimental framework adopted for refining an Italian Large Language Model (LLM) by leveraging a novel corpus generation technique introduced in this study. The experiments are based on the state-of-the-art 7-billion parameter model, \texttt{mistral-7b}, which currently stands as the most proficient model within its size category \cite{mistral7b}

The primary objective of the following experiments is to gain a deeper understanding of how the choice of corpus impacts the performance and Italian proficiency of the fine-tuned model. To explore this, we subject the \texttt{mistral-7b} model to three distinct fine-tuning conditions. First, it undergoes fine-tuning exclusively with the \textit{Fauno} corpus. Second, it undergoes fine-tuning exclusively with the \textit{Generated} corpus. Finally, we conduct a fine-tuning process using a combination of the \textit{Fauno} corpus and our newly generated corpus. This allows us to analyze the effects of different corpus compositions on benchmark performance.

It is important to notice that before generating the final fine-tuning corpora, both the \textit{Generated} Dataset, \textit{Fauno}, and \textit{Full} were filtered to only retain messages with a quality score below 2. This pre-processing step was crucial to ensure the quality and relevance of the training data, impacting the subsequent fine-tuning efficiency and efficacy of the model.

The evaluation of the fine-tuned models will involve assessing their performance in comparison to two well-established Italian language models, namely, \textit{Camoscio} and \textit{Fauno}, as well as the base model \texttt{mistral-7b}. Initially, this comparison will focus on a few-shot question-answering task using the Italian version of the SQuAD dataset. Additionally, we will evaluate the performance of these models across three other few-shot tasks taken from the EVALITA benchmark, which include toxicity detection, irony detection, and sentiment analysis. These tasks are crucial for evaluating the models' nuanced understanding of language and contextual interpretation, making them essential for a comprehensive assessment of model performance.

\subsection{Training Environment and Parameters}

The training of our models was conducted in a modern computational environment, utilizing an NVIDIA DGX system equipped with 8 H100 GPUs. This high-performance setup provided the necessary computational power and efficiency to handle the extensive training demands of our Large Language Models (LLMs). The consistent use of this advanced hardware configuration across all our experimental trials ensured uniformity and comparability in the training process. Each model, regardless of the dataset it was fine-tuned with, underwent training under identical conditions, leveraging the full capabilities of the DGX system's parallel processing. 

The chosen hyperparameters are detailed in Table \ref{tab:hyperparameters}, these parameters included the base model configuration, model and tokenizer types, sequence length, batch size, number of epochs, and learning rate, among others. The optimizer was set to \texttt{adamw\_bnb\_8bit}, a variant of the AdamW optimizer, known for its effectiveness in training large neural networks. Additionally, the learning rate was set to 0.000005 with a cosine learning rate scheduler, ensuring a gradual and stable learning process. 

Furthermore, the models were trained using \texttt{bfloat16} data type, striking a balance between computational efficiency and numerical precision. The use of Flash Attention \cite{flash-attention},  a feature in H100 GPUs, enabled efficient memory usage and faster computation times, particularly crucial for training models with billions of parameters. The warmup steps were set to 10, allowing the models to gradually adapt to the training regime, thus preventing abrupt changes that could adversely affect the learning process.

This homogeneous training environment and parameter setting across all trials were instrumental in maintaining the consistency of our experimental evaluations. By controlling these variables, we could attribute observed differences in model performance solely to the effects of the fine-tuning datasets, thereby ensuring the validity and reliability of our experimental results.

\begin{table}[h!]
\centering
\caption{Hyperparameters for Model Training}
\label{tab:hyperparameters}
\begin{tabular}{ll}
\hline
\textbf{Hyperparameter}                & \textbf{Value}                       \\ \hline
Base Model                             & \texttt{mistralai/Mistral-7B-v0.1}   \\ 
Model Type                             & \texttt{MistralForCausalLM}          \\ 
Tokenizer Type                         & \texttt{LlamaTokenizer}              \\ 
Sequence Length                        & 8192                                 \\ 
Micro Batch Size                       & 4                                    \\ 
Number of Epochs                       & 1                                    \\ 
Optimizer                              & \texttt{adamw\_bnb\_8bit}            \\ 
Learning Rate                          & 0.000005                             \\ 
LR Scheduler                           & \texttt{cosine}                      \\ 
dtype                                  & \texttt{bfloat16}                    \\ 
Flash Attention                        & Yes                                  \\ 
Warmup Steps                           & 10                                   \\ \hline
\end{tabular}
\end{table}

\subsection{Benchmarks}

\begin{table}[h]
\centering
\caption{Few-shot F1 scores and Exact Match (EM) on the SQuAD-it dataset}
\label{tab:squad_it_results}
\begin{tabular}{lccc}
\hline
\textbf{Model} & \textbf{F1} & \textbf{EM} \\
\hline
\textbf{\texttt{mistral-7b} f.t.w. \textit{Full}} \textbf{(\texttt{cerbero-7b})}                 & \textbf{72.55\%} & \textbf{55.6\%} \\
\textbf{\texttt{mistral-7b} f.t.w. \textit{Generated}}            & \textbf{70.83\%} & \textbf{52.5\%} \\
\texttt{mistral-7b} f.t.w. \textit{Fauno}                         & 68.35\% & 48.2\% \\
\hline
Base \texttt{mistral-7b}                                          & 15.55\% & 8.50\% \\
\hline
\texttt{Fauno}                                                    & 44.46\% & 0.00\% \\
\texttt{Camoscio}                                                 & 37.42\% & 0.00\% \\
\hline
\end{tabular}
\end{table}

\begin{table*}[h]
\centering
\caption{F1 scores for few-shot tasks from the EVALITA benchmark}
\label{tab:evalita_results}
\begin{tabular}{lccc}
\hline
\textbf{Model} & \textbf{Sentiment Analysis} & \textbf{Irony Detection} & \textbf{Toxicity} \\
\hline
\textbf{\texttt{mistral-7b} f.t.w. \textit{Full}} \textbf{(\texttt{cerbero-7b})}              & \textbf{61.80\%} & \textbf{48.51\%} & \textbf{63.04\%} \\
\textbf{\texttt{mistral-7b} f.t.w. \textit{Generated}}        & \textbf{59.72\%} & \textbf{44.75\%} & \textbf{63.74\%} \\
\texttt{mistral-7b} f.t.w. \textit{Fauno}                     & 57.45\%          & 41.84\%          & 63.46\% \\
\hline
Base \texttt{mistral-7b}                                      & 12.14\%          & 34.16\%          & 34.16\% \\
\hline
\texttt{Fauno}                                                & 12.23\%          & 39.17\%          & 33.84\% \\
\texttt{Camoscio}                                             & 13.33\%          & 39.65\%          & 38.18\% \\
\hline
\end{tabular}
\end{table*}

The Stanford Question Answering Dataset (SQuAD) is a reading comprehension dataset consisting of questions posed by crowdworkers on a set of Wikipedia articles. Each question's answer is a segment of text from the corresponding reading passage. It is one of the largest datasets of its kind with 100,000+ question-answer pairs on 500+ articles. SQuAD is known for its diversity and size, making it more comprehensive than previous datasets. In our experimentation, we used the Italian version of SQuAD known as SQuAD-it. SQuAD-it was created by performing a semi-automated translation of the original SQuAD dataset into Italian. This extensive resource comprises over 60,000 question-answer pairs, all geared towards facilitating open question-answering processes in the Italian language. We specifically adopted a three-shot setting for our few-shot question-answering benchmark. By using a three-shot setting, we assess a model's generalization capability, evaluating its capacity to understand and reason across various topics and contexts based on a small set of examples. Moreover, we opted for a three-shot setting in our benchmark to effectively manage the length of context, especially as the question-answer pairs often involve extensive details. This choice strikes a balance between evaluating a model's ability to handle longer information and maintaining a manageable evaluation process.

The EVALITA benchmark is an initiative by the Italian Association for Computational Linguistics (AILC) started in 2007, focusing on the evaluation of Natural Language Processing (NLP) tools for Italian. It operates under a shared framework where diverse systems and methodologies can be assessed across a variety of tasks. These tasks are meant to represent scientific challenges and are organized by the Italian research community to test methods, resources, and systems against shared benchmarks. These benchmarks are indicative of linguistic open issues or real-world applications. EVALITA not only aims at the advancement of methodologies and techniques for NLP and speech processing but also emphasizes reproducibility, cross-community engagement, and the exploration of multilingual and multi-modal dimensions. For EVALITA, we employed a five-shot setting in our evaluations due to the relatively compact nature of both input queries and expected output responses.

From the EVALITA corpus, we selected three benchmarks corpus from which we constructed our few-shot classification tasks: i) AMI ii) IronITA iii) SENTIPOLC:

The AMI 2020 dataset is a valuable resource consisting of 6,000 tweets written in Italian, meticulously annotated for misogyny and aggressiveness. This dataset was employed in the AMI shared task during the Evalita 2020 evaluation campaign, providing a crucial evaluation ground for models in understanding and identifying potentially harmful content in Italian social media interactions.

IronITA, another pivotal benchmark, collects 4,849 tweets that have been expertly annotated for irony and sarcasm. It was utilized in the IroniTA task organized as part of EVALITA 2018, facilitating the assessment of models' ability to detect nuanced language phenomena like irony and sarcasm, which often require a sophisticated understanding of context and sentiment.

The SENTIPOLC dataset, developed for EVALITA 2014, focuses on sentiment polarity classification. It includes short social media posts, primarily from Twitter, annotated to identify whether they express positive, negative, neutral, or mixed sentiments. This dataset uniquely incorporates ironic messages to investigate the challenges of correctly classifying sentiment in such contexts.

\subsection{Results Discussion}

This section delves into the interpretation and implications of the experimental results obtained from the fine-tuning and evaluation of the \texttt{mistral-7b} model under various conditions. The discussion is structured around the performance metrics observed in the SQuAD-it and EVALITA benchmark tasks, with a focus on the few-shot learning capabilities of the models.

The SQuAD-it results, as detailed in Table \ref{tab:squad_it_results}, highlight significant differences in model performance based on the training dataset used. Notably, the \texttt{mistral-7b} model fine-tuned with the \textit{Full} dataset, which includes our generated dataset combined with \textit{Fauno}, outperforms the other configurations in both F1 and Exact Match (EM) scores. This suggests that the richness and diversity of the \textit{Full} corpus contribute positively to the model's comprehension abilities and its precision in answering questions.

The \textit{Generated} corpus, while slightly lower in performance compared to the \textit{Full} corpus, still shows a marked improvement over the \textit{Fauno} corpus. This indicates the effectiveness of our novel corpus generation technique in enhancing the model's language understanding, especially considering the structural assertions and rule-based NLP techniques employed for refinement.

In the EVALITA benchmark, as illustrated in Table \ref{tab:evalita_results}, the few-shot task results offer a nuanced view of the models' capabilities in understanding complex language constructs such as irony, sentiment, and toxicity. The \texttt{mistral-7b} model fine-tuned with the \textit{Full} corpus demonstrates superior performance across all tasks, underscoring the value of a diverse and comprehensive training corpus.

Interestingly, the \textit{Generated} corpus shows comparable or slightly better performance in toxicity detection but falls short in irony and sentiment analysis. This could be attributed to the inherent challenges in generating corpora that adequately capture these subtle aspects of language, emphasizing the need for further refinement in corpus generation methodologies.

When compared against baseline models like the base \texttt{mistral-7b}, \textit{Fauno}, and \textit{Camoscio}, the fine-tuned models exhibit substantial improvements. This reinforces the significance of fine-tuning with task-specific corpora in enhancing model performance, particularly in language-specific contexts like Italian.

The results of this study have several implications for the field of NLP, particularly in the development of LLMs for languages other than English. The effectiveness of the corpus generation and refinement techniques presented here offers a promising avenue for improving language models, especially for underrepresented languages.

Future research could explore further advancements in corpus generation methods, perhaps integrating more sophisticated techniques to capture complex language nuances. Additionally, extending this methodology to other languages and dialects could significantly contribute to the global inclusivity and accessibility of language technologies.

\section{Conclusions}

This research has advanced the field of Natural Language Processing (NLP), with a focus on the generation and refinement of corpora for fine-tuning Large Language Models (LLMs). Our work introduced a novel method for creating a high-quality, diverse scope-specific chat corpus through a sentence transformers-aided self-chat mechanism.

We successfully generated a new Italian chat corpus and refined the existing \textit{Fauno} corpus, which was substantial in size but lacking in quality. Our innovative approach combined self-chat mechanisms with structural assertions and NLP rules to enhance the quality and diversity of the corpora. Furthermore, we introduced a novel Masked Language Modelling (MLM) model-based metric to assess and compare the quality of our generated corpus against existing corpora. This metric has proven to be a reliable method for evaluating language model corpora.

Fine-tuning the Italian LLM with our newly generated corpus, both in isolation and combined with the \textit{Fauno} corpus, led to significant improvements in language comprehension and question-answering capabilities. Our benchmarks, conducted using the SQuAD-it and EVALITA tasks, demonstrated the enhanced performance of the fine-tuned models, especially in understanding complex constructs such as irony, sentiment, and toxicity.

The implications of our research are far-reaching within the domain of NLP. We underscore the potential of advanced corpus generation and refinement techniques in improving the performance of LLMs, particularly for underrepresented languages like Italian. This advancement is crucial for global language inclusivity in AI applications. Our novel methodologies in corpus generation, particularly the use of self-chat mechanisms complemented by embedder LLMs, set a new standard in corpus creation and can be applied to other languages. The MLM-based quality assessment metric we developed provides a new benchmark for evaluating language corpora and can be utilized in future studies.

While our study has achieved significant milestones, it also opens avenues for future research. Extending our corpus generation and refinement methodology to other languages and dialects would contribute significantly to the diversity and inclusivity in the field of conversational AI. There is also scope for further refinement of our corpus generation techniques, especially in capturing more nuanced aspects of language. 

In conclusion, our research presents a significant leap forward in the generation and refinement of language corpora, particularly for Italian LLMs. The methodologies developed and the insights gained lay a solid foundation for future endeavors in the NLP domain, paving the way for more inclusive and effective conversational AI systems.

To foster collaboration and advance Natural Language Processing (NLP), we've publicly released our model and corpus on GitHub and Hugging Face \cite{cerbero7b}.  This release aims to inspire further research, and innovation in fine-tuning corpora and large language models, and promote transparency and inclusivity in AI. We encourage the NLP community to build upon our work, expand it to other languages, and contribute to the development of more capable and culturally diverse language models.

\section*{Acknowledgments}

Work partially supported by: (i) the University of Pisa, in the framework of the PRA 2022 101 project ``Decision Support Systems for territorial networks for managing ecosystem services''; (ii) the European Commission under the NextGenerationEU program, Partenariato Esteso PNRR PE1 - ``FAIR - Future Artificial Intelligence Research'' - Spoke 1 ``Human-centered AI''; (iii) the Italian Ministry of Education and Research (MIUR) in the framework of the FoReLab project (Departments of Excellence), in the framework of the ``Reasoning'' project, PRIN 2020 LS Programme, Project number 2493 04-11-2021, and in the framework of the project ``OCAX -Oral CAncer eXplained by DL-enhanced case-based classification'' PRIN 2022 code P2022KMWX3. Work partly funded by the European Commission under the NextGeneration EU program, PNRR - M4 C2, Investment 1.5 ``Creating and strengthening of innovation ecosystems'', building ``territorial R\&D leaders'', project ``THE - Tuscany Health Ecosystem'', Spoke 6 ``Precision Medicine and Personalized Healthcare''.

\printbibliography

\begin{IEEEbiography}[{\includegraphics[width=1in,height=1.25in,clip,keepaspectratio]{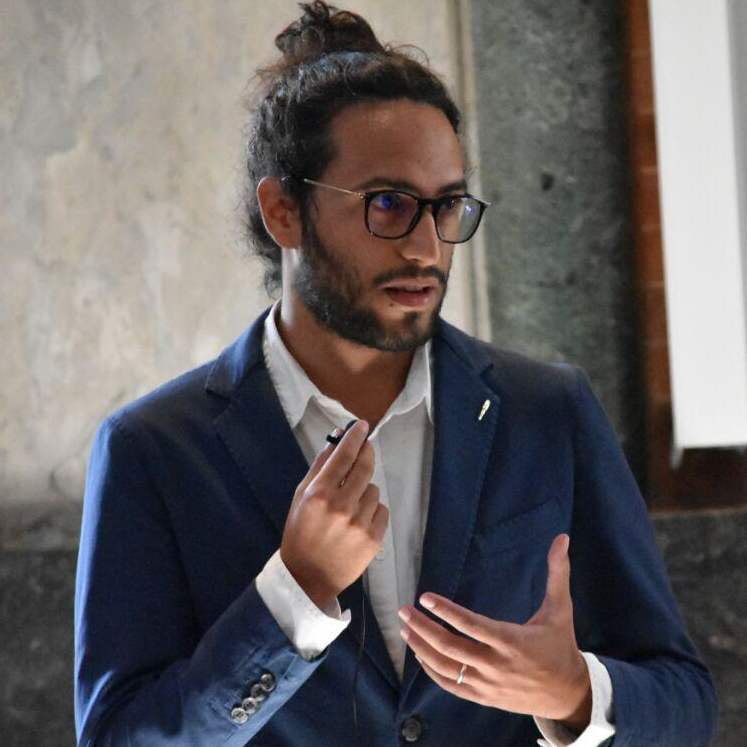}}]{Federico A. Galatolo}
is an Assistant Professor at the Department of Information Engineering of the University of Pisa (Italy). His research expertise spans the domains of Deep Learning, Artificial Intelligence, and their applications across various fields. He has contributed significantly to the academic community as a (co-)author of numerous international scientific publications. He is also a prolific contributor to the open-source community, having developed and maintained over 100 public repositories.\end{IEEEbiography}

\begin{IEEEbiography}[{\includegraphics[width=1in,height=1.25in,clip,keepaspectratio]{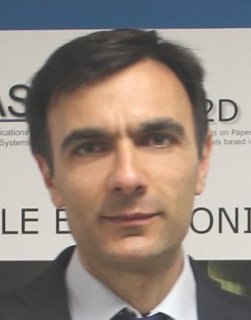}}]{Mario G.C.A. Cimino}
is an Associate Professor at the Department of Information Engineering of the University of Pisa (Italy). His research lies in the areas of Information Systems and Artificial intelligence. He is (co-)author of about 100 international scientific publications. He is an Associate Editor of the Journal of Granular Computing (Springer) and the Journal of Ambient Intelligence and Humanized Computing (Springer). He is Vice-Chair of the IEEE CIS Task Force "Intelligent Agents", IEEE Computational Intelligence Society.\end{IEEEbiography}

\end{document}